\documentclass[twoside,leqno,twocolumn]{article}
\usepackage{ltexpprt}

\usepackage{etoolbox}
\AtBeginEnvironment{tabular}{\small}

\usepackage{times}
\usepackage{soul}
\usepackage[small]{caption}

\usepackage[table]{xcolor}
\usepackage[utf8]{inputenc} 
\usepackage{hyperref}       
\usepackage{url}            
\usepackage{booktabs}       
\usepackage{amsfonts}       
\usepackage{amssymb}
\usepackage{dsfont}
\usepackage[noend]{algorithmic}
\usepackage{algorithm}
\usepackage{amstext}
\usepackage{graphicx}
\usepackage{cuted}

\usepackage{multirow}
\usepackage{amsmath}
\usepackage{cuted}
\usepackage{arydshln} 
\usepackage{dashrule}

\usepackage{caption} 
\usepackage{subcaption}

\usepackage{threeparttable,booktabs}
\usepackage{array}
\newcolumntype{@}{>{\global\let\currentrowstyle\relax}}
\newcolumntype{^}{>{\currentrowstyle}}
\newcommand{\rowstyle}[1]{\gdef\currentrowstyle{#1}%
  #1\ignorespaces
}
\usepackage{xcolor}
\definecolor{hcolor}{rgb}{0.90,0.90,0.90}
\definecolor{hcolor1}{rgb}{0.90,0.90,0.90}
\definecolor{hcolor2}{rgb}{0.94,0.94,0.94}
\definecolor{hcolor3}{rgb}{0.98,0.98,0.98}

\title{Improving the Inference of Topic Models \\ via Infinite Latent State Replications}

\iftrue
\author{
Daniel Rugeles$^1$, 
Zhen Hai$^1$, 
Juan Felipe Carmona $^2$
Manoranjan Dash $^3$, 
Gao Cong$^1$,
\\ 
$^1$ School of Computer Science and Engineering, Nanyang Technological University \\
$^2$ Departamento de Matem\'aticas, Universidad de los Andes, Colombia\\
$^3$ School of Computing, National University of Singapore\\
\{daniel007, haiz0001\}@e.ntu.edu.sg,
jf.carmona658@uniandes.edu.co,
mano@comp.nus.edu.sg,
gaocong@ntu.edu.sg
}
\fi

\begin{document}

\maketitle

\begin{abstract}
In text mining, 
topic models are a type of probabilistic generative models for inferring latent semantic topics from text corpus. 
One of the most popular inference approaches to topic models is perhaps collapsed Gibbs sampling (CGS), 
which typically samples one single topic label for each observed document-word pair.
In this paper, 
we aim at improving the inference of CGS for topic models.
We propose to leverage state augmentation technique by maximizing the number of topic samples to \textit{infinity},
and then develop a new inference approach, called infinite latent state replication (ILR), to generate robust soft topic assignment for each given document-word pair.
Experimental results on the publicly available datasets show that 
ILR outperforms CGS for inference of existing established topic models.


\end{abstract}

\section{Introduction}
\label{submission}

In text mining, 
probabilistic topic models, 
such as latent Dirichlet allocation (LDA) \cite{lda}, 
refer to generative statistical algorithms for mining latent semantic structure of a set of text documents. 
Aside from their increasing popularity and ubiquitous adoption in text data processing, 
they have been also applied to the fields such as computer vision \cite{visiontopic} and geographical modeling \cite{geotopic}. 
Recently, 
widespread applicability has spurred research on developing more accurate or efficient inference approaches to topic modeling.


Existing inference methods can be roughly grouped into \textit{optimization} and \textit{sampling} categories. 
One of the representative optimization based approaches is the mean field variational inference (VI) \cite{reviewvi}. 
VI frames the inference process as an optimization problem, 
where the posterior is approximated by fitting a selected family of distributions. 
Unfortunately, 
the predictive ability of VI is somewhat sacrificed for computational efficiency \cite{VIMCMC}.
On the other hand, 
one of the most popular sampling based inference approaches is perhaps the collapsed Gibbs sampling (CGS). 
CGS has been shown effective and widely used to infer latent topic models \cite{ldacgs}.
In this work, 
our focus is on improving the generalization performance of CGS for inference of topic models.






Particularly, 
in the inference of CGS for LDA, 
for each observed document-word pair, 
it estimates a posterior distribution conditioned on the topic assignments of all the other pairs,
and relies on the distribution to sample one single topic for the given pair. 
The model's parameters can be then estimated based on the sample data.
Intuitively, 
instead of sampling one exact topic assignment,
we may yield a flexible and more robust learning approach 
by sampling multiple topics 
and then combining the topic samples to derive \textit{soft} topic assignment for each document-word pair.




Considering a basic situation 
where two samples are drawn for the given pair,
the two samples correspond to two topic assignments,
and combining the two samples would lead to flexible search for optimal assignment by using the two branches simultaneously. 
Then, 
as the number of samples increases, 
the parameter estimation would be done 
by relying on more and more evidence from all of the samples. 
This process is known as state augmentation for marginal estimation. 
It was previously studied for improving maximum a posteriori estimation 
using Markov Chain Monte Carlo method (MCMC) \cite{MAPMCMC}.

We extend the idea, 
and propose to leverage the benefit of maximizing the state augmentation 
by raising the number of samples to \textit{infinity}.
Then,
we develop a new method, called infinite latent state replication (ILR), to improve the inference of topic models within Gibbs sampling framework.
As a matter of fact,
the computational complexity of the inference model 
often increases linearly with the number of samples. 
To deal with this issue,
we employ the strong law of large numbers, 
and present a tractable optimization solution 
that achieves the same computational complexity of CGS.


In contrast to collapsed Gibbs sampling,
ILR leverages the aggregation of whole population to estimate the density of the posterior distribution. 
Thus, 
sampling is not required in ILR, 
and the resulting inference algorithm is deterministic.
Thanks to the determinism of ILR, 
its convergence can be assessed directly by checking when the parameters of the posterior distribution stop changing. 
In contrast, 
the exact iteration at which CGS converges is unknown, 
and additional mechanisms may be required to assess its convergence.

Next,
we apply ILR to inference of LDA, 
and observe a significant improvement over CGS in predictive perplexity. 
In addition,
ILR can be also applied to inference of other topic models 
that characterize the dependency between two latent random variables,
e.g., dual topic model (DTM) \cite{dt2b}. 

This work has made the following main contributions:
\begin{itemize}
\item Based on Gibbs sampling framework, 
we develop ILR to estimate the posterior for inference of topic models. 
ILR replaces the exact assignments of latent topics 
by a density given by drawing multiple topic assignments.
It is thus a less constrained learning method compared to CGS, 
and may have flexibility and better chance to improve the generalization performance for inference.
\vspace{-0.2cm}
\item The replication of topic assignments often results in linear increase in computational complexity. 
To tackle this issue,
we maximize the number of replications to infinity, 
and transform the Gibbs sampling algorithm into a deterministic method, 
whose computational complexity is equivalent to standard Gibbs sampling algorithm.
\vspace{-0.2cm}
\item We conduct extensive experiments on the publicly available benchmark datasets,
and experimental results validate the improved effectiveness of ILR over CGS for inference of existing well-established topic models. 
\end{itemize}

\section{Related Work}

Along the rich history of Bayesian models, several inference algorithms have been proposed, such as variational Bayesian inference \cite{lda}, expectation propagation \cite{ldaep}, collapsed Gibbs sampling \cite{ldacgs}, and belief propagation \cite{ldabp}. Among these algorithms, collapsed Gibbs sampling and variational inference are perhaps the most widely used methods due to their effectiveness. The former, which employs a sampling strategy, is known for guaranteeing the convergence to true target posterior, while the latter, which is a theoretically backed optimization method, approximates the true posterior.

As far as we know, 
existing extensions on CGS are primarily concerned with improving runtime. 
This is natural given that CGS tends to result in better predictive performance than other inference methods, 
but it is computationally expensive. 
FastLDA was shown to be eight times faster than CGS while maintaining the comparable performance \cite{fastlda}. 
The high computational cost of CGS stems from the $O(K)$ time complexity incurred at every sampling step ($K$: topic number). 
It is observed that, 
even given a good many topics, 
words or especially infrequent words are often assigned only to a few topics. 
Then, by keeping track of these words, FastLDA requires significantly less than $K$ operations per sample on average. 
SparseLDA factorizes the posterior equation into a sum of three factors. 
The sampling scheme is then replaced by a uniform sampling strategy, where the probability mass falls in one of the buckets in 90\% of the time \cite{sparsefastlda}. 
By using an appropriate data structure, 
the computation for the mass can be optimized. 
As a result, the model attains an order of magnitude improvement in computational complexity 
without affecting its predictive power compared to CGS. Recently, AliasLDA \cite{aliaslda} and LightLDA \cite{LightLDA} are able to reduce the $O(K)$ complexity of sampling by an efficient sampling method from discrete random variables. In the latter case, a Metropolis-Hasting step allows sampling a topic assignment for a document-word pair in $O(1)$ via Walker's alias method \cite{aliastable}. Then, WarpLDA improved LightLDA by exploiting the memory access behavior of other algorithms for LDA \cite{WARPLDA}.

In contrast to sampling-based inference methods, 
existing work on variational inference approaches mainly aims to improve the predictive performance. 
One reason which may illustrate the weakness of these methods perhaps lies in the approximation to intractable integrals.  
Collapsed variational Bayesian schemes \cite{ldacvb} use a second-order Taylor expansion to approximate the integrals. 
In addition,
the zero-order information can be also used in the variational methods,
and may result in improved inference \cite{cvb0}.
 



In this work, 
rather than targeting the efficiency of CGS, 
we aim to improve the predictive ability of CGS for inference of latent topic models. 
To achieve that, 
we propose to leverage a state augmentation technique, 
which normally augments the probabilistic model 
by replicating multiple times the latent state random variables.
The technique has been previously shown effective for improving 
the marginal maximum a posteriori estimation via the Markov chain Monte Carlo \cite{MAPMCMC}. 
Moreover,
we show that it is possible to maximize the benefit of the state augmentation technique by raising the number of replications up to infinity. 
In this way, 
we introduce a new deterministic algorithm to tackle the inference of topic models.
Our method can lead to improved predictive performance,
while it maintains equivalent computational complexity compared to the standard collapsed Gibbs sampling algorithm.



\section{Preliminaries}

\label{sec:preliminaries}

Latent Dirichlet allocation (LDA), one of the most common topic models, 
is able to discover $K$ topics that pervade an unstructured collection of documents. 
In LDA, we assume that each document has a multinomial distribution over $K$ topics parameterized by $\theta$. 
We use $\theta_{d}$ as the set of parameters for representing the probability distribution over the set of topics given a document $d$.
Each topic is defined as a multinomial distribution over the vocabulary of words parameterized by $\phi$. 
We use $\phi_{k}$ as the set of parameters of 
the probability over the set of words given a topic $k$. 
Per document topic distribution $\theta_d$ is generated by sampling from a Dirichlet distribution with hyper-parameter $\alpha$. Similarly, $\phi_k$ is generated by sampling from a Dirichlet distribution with hyper-parameter $\beta$.

In latent Dirichlet allocation via collapsed Gibbs sampling (LDA-CGS), 
to find latent topics, 
we use a dataset $X$ that consists of $M$ observed document-word pairs $x_j=(d_j,w_j), ~\forall j \in 1 \cdots M$. 
where $|D|$, and $|W|$ are the number of documents and the vocabulary size.
Each pair is assigned exactly to one of the topics as established by the assumption of the model. 
For each $j^{th}$ pair,
the topic assignment is generated by sampling the conditional posterior distribution of $Z_j$ given the topic assignments of all other pairs $Z_{-j}=z_{-j}$. 
Formally, 
we represent the conditional posterior distribution as $p(Z_j|Z_{-j}=z_{-j},\textbf{X})$. 
The probability density of the posterior distribution is given in Equation \ref{eq:ldacgs}.

\vspace{-0.5cm}
\begin{equation}
\label{eq:ldacgs}
f(Z_j|Z_{-j}=z_{-j}) \propto \frac{(D_{d_j}+ \alpha) (W_{w_j}+\beta)}{(N_j+V\beta)}
\end{equation}

We use the subscript $Z_j|Z_{-j}$ to denote the density of the posterior distribution $p(Z_j|Z_{-j}, \textbf{X})$. 
The $D_{d_j}, W_{w_j}$, and $N_j$ are count vectors computed using the indicator function $\mathds{1}()$:
{\small
\begin{gather}
D_{d_j}={\sum_{\underset{\large m\neq j}{m=1}}^{M}}\mathds{1}(d_j=d_m)*\bar{z_m} ~;~ W_{w_j}={\sum_{\underset{\large m\neq j}{m=1}}^{M}}\mathds{1}(w_j=w_m)*\bar{z_m} \nonumber\\
N_j={\sum_{\underset{\large m\neq j}{m=1}}^{M}}\bar{z_m}
\label{eq:factors}
\end{gather}
}


\noindent
where $\bar{z_m}$ refers to the realization of the random variable $Z_m$. 
We use the bar notation to represent a one-hot encoding representation. 
Hence, $D_{d_j}$ denotes the vector of topic assignments for document $d_j$, $W_{w_j}$ denotes the vector of topic assignments for word token $w_j$, 
and $N_j$ represents the vector of topic assignments over the entire set of document-word pairs. 
This representation facilitates the comparison of Equation \ref{eq:ldacgs} with the posterior density proposed in Section \ref{sec:ILR}.

The collapsed Gibbs sampling approach to LDA (LDA-CGS) is summarized in Algorithm \ref{alg:cgs}. 
Note that, 
at iteration $i$, 
the computation for the posterior density in Line 5 not only depends on the topic assignments of all the pairs whose index values are smaller than $j$, as denoted by $z^{i}_{<j}$, 
but also depends on the topic assignments of all the pairs with index values larger than $j$ at iteration $i-1$, 
as denoted by $z^{i-1}_{>j}$. 
Thus, we denote the density at the $i^{th}$ iteration for the $j^{th}$ record to be $f(Z^i_j|Z^i_{<j}=z^i_{<j}, Z^{i-1}_{>j}=z^{i-1}_{>j})$.

\begin{algorithm}
   \caption{Collapsed Gibbs Sampling for LDA}
   \label{alg:cgs}
\begin{algorithmic}[1]
   \STATE {\bfseries Input:} Hyper-parameters $K,\alpha,\beta$, dataset $\textbf{X}$
   \STATE Initialize $z^0_j$ using a uniform distribution, and set initial iteration $i=1$
   \REPEAT 
   \FOR{$j=1 \cdots M$}
   \STATE Compute $f(Z^i_j|Z^i_{<j}=z^i_{<j}, Z^{i-1}_{>j}=z^{i-1}_{>j})$ from Equation\ref{eq:ldacgs}
   \STATE Draw $z_j^i$ from $f(Z^i_j|Z^i_{<j}=z^i_{<j}, Z^{i-1}_{>j}=z^{i-1}_{>j})$
   \ENDFOR
   \STATE $i=i+1$
   \UNTIL{Convergence}
   \STATE Collect samples and estimate $\hat{\theta}$ and $\hat{\phi}$ as shown in \cite{ldacgs}
   \RETURN $\hat{\theta},~\hat{\phi}$
\end{algorithmic}
\end{algorithm}


The convergence of LDA-CGS can be obtained 
when the topic assignments $z_j$ for each pair are drawn from the marginal distributions $p(Z_j)$. 
Unfortunately, 
since we do not have an analytical form for $p(Z_j)$, 
we cannot measure exactly at which iteration the samples are drawn from $p(Z_j)$.

\section{Infinite Latent State Replication Inference}
\label{sec:ILR}

When applying collapsed Gibbs sampling to LDA, 
for the $j^{th}$ document-word pair, 
we compute the posterior
$p(Z_j|Z_{-j}=z_{-j})$ by using the samples $z_{-j}$ obtained for all other pairs.
Then, 
we draw \textsc{one} single sample $z_j$ from this distribution, 
and continue to compute the posterior for the next document-word pair $p(Z_{j+1}|Z_{-(j+1)}=z_{-(j+1)})$. 
In collapsed Gibbs sampling, 
the posterior distributions are updated until the convergence is reached.
Table \ref{tab:cgs} shows a toy example of applying collapsed Gibbs sampling to inference of LDA, 
where the number of topics is $K=3$.
For each of $M$ observed document-word pairs $(d_j, w_j)$,
a topic label ($1, 2,$ or $3$) is assigned by relying on the posterior probability distribution (last column). 


\begin{table}[ht]
  \caption{A toy example of applying collapsed Gibbs sampling to inference of LDA ($K=3$)}
  \label{tab:cgs}
  \centering
   \small
  \begin{tabular}{cccc} 
    \toprule
    $j$ & $(d_j, w_j)$ & $z_j$ & Posterior prob. \\
    \midrule
    1 & $(d_1, w_1)$  & 2 & [ 0.2 0.3 0.5 ]\\
    $\vdots$ & $\vdots$  & $\vdots$  & $\vdots$ \\
    M & $(d_M, w_M)$  & 2 & [ 0.2 0.4 0.4]\\
    \bottomrule
  \end{tabular}
\end{table}



In contrast,
the replication of latent states augments the collapsed Gibbs sampling algorithm by repeatedly drawing \textsc{R} samples from each posterior distribution to compute the newly updated posterior distributions. 
The replication of latent space has been previously studied to improve the parameter estimate in maximum a posteriori estimation 
by using Markov Chain Monte Carlo estimations (MCMC). 
In MCMC, 
the replications of the latent space actually provide more evidence about the search path, 
which can lead to a more robust exploration of the parameter space \cite{MAPMCMC}.


To our knowledge,
the replication of latent states does not affect the global optima of the original model. 
The augmented model with $R$ replications described by $p'(\Theta,Z^{(1)}, .. Z^{(R)},X)$ is the $R^{th}$ power of the model that does not use replications \cite{SAME}, as shown below. 

\vspace{-0.6cm}
\begin{equation}
p'(\Theta,Z^{(1)}, .. Z^{(R)},X) \propto \prod_{r=1}^R p(\Theta,Z^{(r)},X)
\end{equation}

\vspace{-0.2cm}
In latent replication inference,
the replications are defined as multiple ($R$) copies of the latent variables:

\vspace{-0.6cm}
\begin{gather}
Z_j^{(r)} =\dfrac{1}{R} Z_j, \forall j \in 1 \cdots M, \forall r \in 1 \cdots R
\end{gather}

A linear transformation can be adopted to maintain the expected value of the random variables $Z_j, \forall j \in 1 \cdots M$, without affecting the optima of the model. 
However, 
this transformation is limited to models,
where the latent random variable follows a multinomial distribution.


\begin{table}[ht]
  \caption{A toy example of applying $R=4$ replications in collapsed Gibbs sampling to inference of LDA ($K=3$). The topic assignment is performed repeatedly for $R$ times for each observed document-word pair (separated by dotted lines).}
  \label{tab:same}
  \centering
  \small
  \begin{tabular}{ c c c c}
    \toprule
    $j$ & $(d_{j}, w_{j})$ & $z^{(r)}_j$ & Posterior prob. (Equation \ref{eq:post1}) \\
    \midrule
   1 & $(d_1, w_1)$  & 2 & [ 0.22 0.28 0.5 ]\\
    1 & $(d_1, w_1)$  & 3 & [ 0.22 0.28 0.5 ]\\
    1 & $(d_1, w_1)$  & 1 & [ 0.22 0.28 0.5 ]\\
    1 & $(d_1, w_1)$  & 3 & [ 0.22 0.28 0.5 ]\\
    \hdashline
    $\vdots$ & $\vdots$ & $\vdots$ & $\vdots$ \\
    \hdashline
   M & $(d_M, w_M)$   & 2 & [ 0.2 0.39 0.41]\\
    M & $(d_M, w_M)$ & 3 & [ 0.2 0.39 0.41]\\
   M & $(d_M, w_M)$   & 2 & [ 0.2 0.39 0.41]\\
    M & $(d_M, w_M)$   & 3 & [ 0.2 0.39 0.41]\\
    \bottomrule
  \end{tabular}
\end{table}

We then employ multiple replications of latent states in collapsed Gibbs sampling 
for inference of LDA. 
Formally, 
in CGS, we draw one single sample from each posterior, 
and then the computation of the $j^{th}$ posterior, 
$p(Z_j|Z_{-j}{=}z_{-j}), \forall j \in 1 .. M$, 
relies on \textsc{one} point estimate from each of the other posteriors,
i.e., $z_i, \forall i \in 1 .. M, i \neq j$. 
In contrast,
in latent state replication inference,
we draw multiple ($R$) samples from each posterior, 
and then the computation of the $j^{th}$ posterior would leverage \textsc{multiple} point estimates from other posteriors,
i.e., $z^{(r)}_i, \forall i \in 1 .. M, \forall r \in R, \and i \neq j$. 
Therefore, 
the combination of multiple point estimates for a posterior serves to approximate the density of the posterior. 
Then,
taking into account the replicas, the $j^{th}$ posterior corresponds to:
$P(Z_j|Z_{-j}^{(1)}{=}z_{-j}^{(1)}, \cdots, Z_{-j}^{(R)}{=}z_{-j}^{(R)} )$, 
where $R$ is the number of replications for estimates.

Table \ref{tab:same} shows a toy example of applying $R=4$ replications of latent states in collapsed Gibbs sampling to inference of LDA ($K=3$).
Conventionally,
for a given document-word pair, e.g., $(d_1, w_1)$,
CGS simply uses one sample, e.g., $z^1_1=2$, as the estimate of the categorical posterior distribution $[0.22 0.28 0.5]$, 
which estimates the posterior using a histogram with proportions given by $[0.00 1.00 0.00]$.
In contrast, 
the state replication inference uses $R=4$ point estimates which can be aggregated in a histogram with proportions given by $[0.25 0.25 0.5]$ (See the first group in Table \ref{tab:same}). 
Clearly, the estimate based on the four replications approximates the posterior $[0.22 0.28 0.5]$ with higher fidelity.
This may suggest that 
the replication of latent states empowers the inference process with a more informative path, which can be exploited by Gibbs chain to converge to a better optima. 
Intuitively, 
while CGS uses a \textit{hard} topic assignment given by the point estimate, 
the state replication inference leverages a \textit{soft} topic assignment,
i.e., splitting the assignment along the $K$ coordinates of the latent space.


\vspace{-0.4cm}
\begin{flalign}
\label{eq:post1}
&p'(Z_j|Z^{(1)}_{-j} .. Z_{-j}^{(R)},X) \propto \int \int   p'(\theta,\phi,Z_{-j}^{(r)} Z^{(R)},X) ~d\theta ~d\phi & \nonumber\\
\vspace{-0.8cm}
& = \prod^{|D|}_m \int 
	\dfrac{\Gamma(\sum_k \alpha_k)}{\prod^{K}_k \Gamma(\alpha_k)}
     \prod^{K}_k \theta^{\alpha_k-1+\frac{1}{R}\sum^R_r c_{k^{(r)},m,*}}_{m,k} d\theta_m \times  &\nonumber \\ 
&\hspace{0.5cm} \prod^{K}_k \int 
	\dfrac{\Gamma(\sum_j \beta_j)}{\prod^{K}_k \Gamma(\beta_k)}
     \prod^{|W|}_j \phi^{\beta_k-1+\frac{1}{R}\sum^R_r c_{k^{(r)},m,*}}_{k,j}  d\phi_k & \nonumber \\ 
\vspace{-0.8cm} 
& =  \prod^{|D|}_m \dfrac{\prod^{K}_k \Gamma( \frac{1}{R}\sum_r c_{k^{(r)},m,*}+\alpha_k)}
              {\Gamma(\sum_k \frac{1}{R}\sum_r c_{k^{(r)},m,*}+\alpha_k)}
\times &\nonumber\\
& \hspace{0.5cm}\prod^{K}_k \dfrac{\prod^{|W|}_j \Gamma( \frac{1}{R}\sum_r c_{k^{(r)},*,j}+\beta_j)}
              {\Gamma(\sum_j \frac{1}{R}\sum_r c_{j^{(r)},*,j}+\beta_j)}  &\nonumber \\
\vspace{-0.8cm} 
& \propto  \dfrac{
	\left( \sum_r c^{-(a,b)}_{z^{(r)}_{a,b},a,*} + \alpha_{z^{(r)}_{a,b}} \right) 
    \times 
    \left( \sum_r c^{-(a,b)}_{z^{(r)}_{a,b},*,y_{a,b}} + \beta_{y_{a,b}}  \right)} 
	{ \sum_r c^{-(a,b)}_{z^{(r)}_{a,b},*,*}+\sum_j \beta_{j}}&
\end{flalign}

\vspace{-0.3cm}

It is worth noting that,
when applying latent state replications within CGS framework, the posterior probability is the same for each replica of a given document-word pair. 
If we employ $R$ latent replications for inference, 
the posterior probability  can be computed 
by following the steps shown in Equation \ref{eq:post1},  some relevant counts are defined as follows. 

\vspace{-0.4cm}
\begin{flalign}
\vspace{-0.2cm}
&c_{k',d',w'} = \sum^M_{m=1} \mathds{1}( z_m=k' \land d_m=d'  \land w_m=w')&\nonumber\\
&c^{-j}_{k',d',w'} = \sum^M_{m=1,m\neq j} \mathds{1}( z_m=k' \land d_m=d'  \land w_m=w')&\nonumber
\end{flalign}

\vspace{-0.2cm}
To facilitate the comparison between standard model and state replication model, 
Equation \ref{eq:post1} may be rewritten in the same form as Equation \ref{eq:factors}:

\vspace{-0.4cm}
{\small
\begin{gather}
D_{d_{j}}={\sum_r^R\sum_{\underset{\large m\neq j}{m=1}}^{M}}\mathds{1}(d_{j}=x^{(d)}_m)*\dfrac{z^{(r)}_m}{R}\\
W_{w_{j}}={\sum_r^R\sum_{\underset{\large m\neq j}{m=1}}^{M}}\mathds{1}(w_{j}=x^{(w)}_m)*\dfrac{z^{(r)}_m}{R} ~;~ N_j={\sum_r^R\sum_{\underset{\large m\neq j}{m=1}}^{M}}\dfrac{z^{(r)}_m}{R} \nonumber
\label{eq:factorsSAME}
\end{gather}
}
\vspace{-0.4cm}


Previous studies have shown that 
the predictive power of applying latent state replication to parameter estimation can be improved with
increasing the number of replications \cite{SAME}, \cite{MAPMCMC}. 
Thus, 
we propose to maximize the number of latent state replications 
to improve the inference of LDA. 
As the number of replications tends toward \textit{infinity}, 
in terms of the law of large numbers, 
we can infer that 
the proportion of the replications obtained by using a categorical distribution will converge to the probability mass of the categorical distribution. 
Mathematically, for all $k \in 1 ..K$: 

\vspace{-0.2cm}
\begin{equation}
\displaystyle\lim_{R \to \infty} \dfrac{1}{R} \displaystyle\sum_{r=1}^R \mathds{1}(z^{r}{=}k) = p'(Z_j=k|Z^{(1)}_{-j} .. Z_{-j}^{(R)},X) = \kappa^k_j,
\end{equation}
where $\kappa^k_j$ is one of the $K$ parameters of the posterior distribution. 
We define  $\kappa_j$ as the vector that holds the parameters of the posterior $\kappa_j= [\kappa^1_j , ..  \kappa^K_j]$. 
Since the posterior is a categorical distribution,
this vector corresponds to the probability mass function of the posterior distribution.

The update of the Gibbs sampler for infinite latent replications 
can be then obtained by raising $R$ to infinity in Equation \ref{eq:factorsSAME}, and the result is shown in Equation \ref{eq:factorsILR}.

\vspace{-0.4cm}
{\small
\begin{gather}
D_{d_{j}}{=}{\sum_{\underset{\large m\neq j}{m=1}}^{M}}\mathds{1}(d_{j}=x^{(d)}_m)*\kappa^k_m ~;~ W_{w_{j}}{=}{\sum_{\underset{\large m\neq j}{m=1}}^{M}}\mathds{1}(w_{j}=x^{(w)}_m)*\kappa^k_m \nonumber\\
N_j={\sum_{\underset{\large m\neq j}{m=1}}^{M}}\kappa^k_m
\label{eq:factorsILR}
\end{gather}
}


Generally, 
from a generative perspective, 
it is computationally unfeasible to sample an infinite number of times. 
As a matter of fact, 
by applying the law of large numbers, 
we can obtain updates that use vector additions without the need for sampling. 
A comparison between Table \ref{tab:cgs} and \ref{tab:same} shows that 
the topic assignment in state replication inference does not correspond to a single label, 
but instead, corresponds to a mixture of labels.


As the number of assigned labels (replications) increases towards infinity, 
the presented updates drive the Gibbs chain 
to convergence by using the whole probability mass from all of the $M$ posterior distributions.
As a consequence, 
the parameters of the inference algorithm are not estimated from samples, but instead are obtained directly by using Equation \ref{eq:factorsSAME}. 
Different from sampling algorithms, 
Equation \ref{eq:factorsSAME} will yield the same result on every iteration after convergence.



The deterministic inference driven by the soft topic assignment mechanism allows for better exploration at the optimal region than the estimations based on random samples. 
As a result, 
ILR achieves significant improvements in generalization performance in terms of predictive perplexity (See Section \ref{sec:exp}).

As for computational cost, 
the parameter updates for each of the document-word pairs has a time complexity of $O(K)$, 
which achieves the same cost as collapsed Gibbs sampler. 
This is actually not a big issue,
as we focus on the improvement of predictive power of CGS under a theoretical framework.



Moreover,
we study the applicability of the proposed infinite latent replication inference to topic models that characterize dependency between latent random variables.


\section{Applicability to Inference of Dual Topic Model}

The dual topic model (DTM) takes as input a co-occurrence data matrix that underlies the inter-relationship between \textit{row} and \textit{column} variables, e.g., user-location matrix where each entry means the number of times a user (row) visits a location (column) \cite{dt2b}. 
Then, it identifies the row topics $Z$ or distribution over the rows of the co-occurrence matrix, the column topics $Y$ or distribution over the columns of the matrix, and the joint distribution of both column and row topics.

The collapsed Gibbs sampling inference for the $j^{th}$ row-column pair is given by Equation \ref{eq:dt2bcgs}:

\vspace{-0.5cm}
\begin{gather}
f(Z_j,Y_j|Z_{-j}{=}z_{-j},Y_{-j}{=}Y_{-j}) \propto \nonumber\\
\frac{(D_{d_{j}}+ \beta^r)^\intercal(W_{w_{j}}+\beta^c)}{(N^r_j+R\beta^r)^\intercal(N^c_j+C\beta^c)}\odot(P_j + \alpha),
\label{eq:dt2bcgs}
\end{gather}
where $R$ and $C$ are the numbers of rows and columns in the input matrix respectively. $\beta^r$ and $\beta^c$ are hyper-parameters used to generate Z and Y respectively, and the following are count vectors computed using the indicator function $\mathds{1}()$:

\vspace{-0.4cm}
{\small
\begin{gather}
D_{d_{j}}={\sum_{\underset{\large m\neq j}{m=1}}^{M}}\mathds{1}(d_{j}=x^{(d)}_m)*\bar{z_m} ~;~
W_{w_{j}}={\sum_{\underset{\large m\neq j}{m=1}}^{M}}\mathds{1}(w_{j}=x^{(w)}_m)*\bar{y_m}\nonumber\\
P_j{=}{\sum_{\underset{\large m\neq j}{m=1}}^{M}}{\bar{z_m}}^\intercal . \bar{y_m}
~;~ N^r_j{=}{\sum_{\underset{\large m\neq j}{m=1}}^{M}}\bar{z_m}  ~;~ N^c_j{=}{\sum_{\underset{\large m\neq j}{m=1}}^{M}} \bar{y_m}
\label{eq:factors_dt2b}
\end{gather}
}
\vspace{-0.4cm}

Note that $\bar{z_m}^T.\bar{y_m}$ is a convenient representation for encoding one unit to a matrix,
where rows have the same dimension as $\bar{z_m}$, 
and the columns have the same dimension as $\bar{y_m}$. 
The $\odot$ refers to element-wise multiplication.

Table \ref{tab:sameDTM}
shows a toy example of applying collapsed Gibbs sampling with latent state replications to inference of DTM,
where the numbers of latent row and column topics are $K^r=2$ and $K^c=2$, respectively.

\begin{table}
  \caption{A toy example of applying collapsed Gibbs sampling with latent state replications to inference of DTM ($K^r=2$, $K^c=2$).}
  \label{tab:sameDTM}
  \centering
    \small
  \begin{tabular}{c c c c >{\footnotesize}c}
    \toprule
   $j$ & $(d_{j}, w_{j})$ & $z^{(r)}_j$ &$z^{(r)}_j$ & Posterior prob. \\
    \midrule
    1 & $(d_1, w_1)$  & 1 & 1 &
    $\left[ \begin{array}{cc} 0.5 & 0  \\ 0 & 0.5 \end{array}\right]$\\
    1 & $(d_1, w_1)$  & 2 & 2 & 
    $\left[ \begin{array}{cc} 0.5 & 0  \\ 0 & 0.5 \end{array}\right]$\\
    1 & $(d_1, w_1)$  & 2 & 2 &
    $\left[ \begin{array}{cc} 0.5 & 0  \\ 0 & 0.5 \end{array}\right]$\\
    1 & $(d_1, w_1)$ & 1 & 1 & 
    $\left[ \begin{array}{cc} 0.5 & 0  \\ 0 & 0.5 \end{array}\right]$\\
    \hdashline
    $\vdots$ & $\vdots$ & $\vdots$ & $\vdots$ \\
    \hdashline
    M & $(d_M, w_M)$ & 1 & 2 & 
    $\left[ \begin{array}{cc} 0.2 & 0.3  \\ 0.4 & 0.1 \end{array}\right]$\\
    M &  $(d_M, w_M)$ & 2 & 1 & 
    $\left[ \begin{array}{cc} 0.2 & 0.3  \\ 0.4 & 0.1 \end{array}\right]$\\
    M &  $(d_M, w_M)$ & 1 & 2 & 
    $\left[ \begin{array}{cc} 0.2 & 0.3  \\ 0.4 & 0.1 \end{array}\right]$\\
    M &  $(d_M, w_M)$ & 1 & 1 & 
    $\left[ \begin{array}{cc} 0.2 & 0.3  \\ 0.4 & 0.1 \end{array}\right]$\\ 
    \bottomrule
  \end{tabular}
\end{table}

According to the law of large numbers, 
the proportion of the joint pairs obtained by the replicas $z^{(r)}_j$  and $z^{(r)}_j$ 
would converge to the value of the parameters of joint posterior probability. For all  $k^r \in 1 .. K^r, \forall k^c \in 1 .. K^c$:

\vspace{-0.5cm}
\begin{gather}
\displaystyle\lim_{R \to \infty} \dfrac{1}{R} \displaystyle\sum_{r=1}^R \mathds{1}(z^{r}{=}k^r,y^{r}{=}k^c) =\\
p(Z_j{=}k^r,Y_j{=}k^c|Z^{(1)}_{-j},Y^{(1)}_{-j} .. Z^{(R)}_{-j},Y^{(R)}_{-j} ){=}\kappa^{k^r,k^c}_j \nonumber
\end{gather}

The $\kappa$ corresponds to a matrix that holds the parameters of the bivariate categorical distribution. 
Following the same procedure as in LDA, we first derive the case for $R$ replications, 
and then drive $R$ up to infinity to derive the ILR inference method applied to DTM.
The resulting factors are shown as follows.

\vspace{-0.2cm}
{\small
\begin{gather}
D_{d_{j}}={\sum_{\underset{\large m\neq j}{m=1}}^{M}}\mathds{1}(d_{j}=x^{(d)}_m)*\kappa_m ~;~ W_{w_{j}}={\sum_{\underset{\large m\neq j}{m=1}}^{M}}\mathds{1}(w_{j}=x^{(w)}_m)*\kappa_m\nonumber\\
P_j{=}{\sum_{\underset{\large m\neq j}{m=1}}^{M}}{\kappa_m} ~;~ N^r_j{=}{\sum_{\underset{m\neq j}{m=1}}^{M}}\kappa_m ~; ~ N^c_j{=}{\sum_{ \underset{\large m\neq j}{m=1}}^{M}}\kappa_m
\label{eq:factors_dt2b_ilr}
\end{gather}
}
\vspace{-0.2cm}

Note that for DTM, the factors are computed from matrix additions in contrast to the vector additions found for LDA.




\section{Experiments}
\label{sec:exp}

\begin{figure*}
\centering
\includegraphics[width=0.33\textwidth,height=3.7cm]{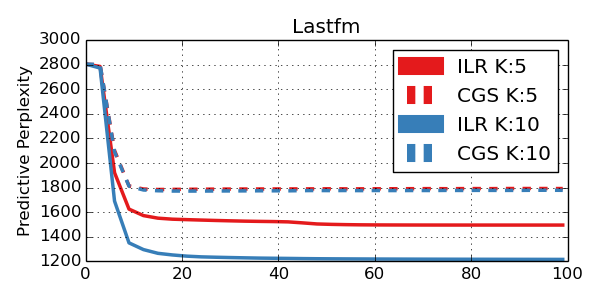}
\includegraphics[width=0.33\textwidth,height=3.7cm]{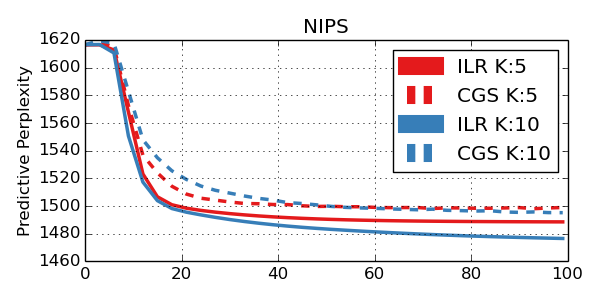}
\includegraphics[width=0.33\textwidth,height=3.7cm]{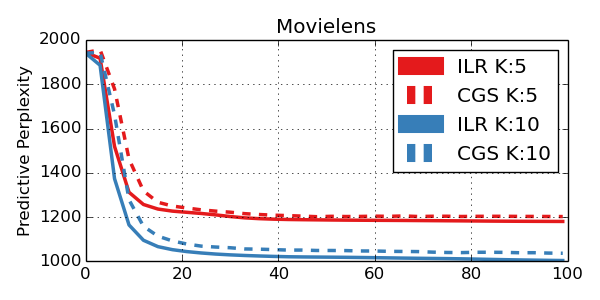}
\includegraphics[width=0.33\textwidth,height=3.7cm]{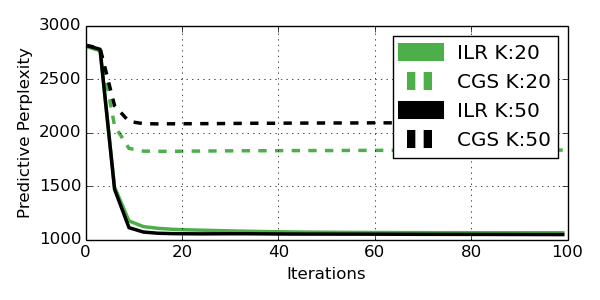}
\includegraphics[width=0.33\textwidth,height=3.7cm]{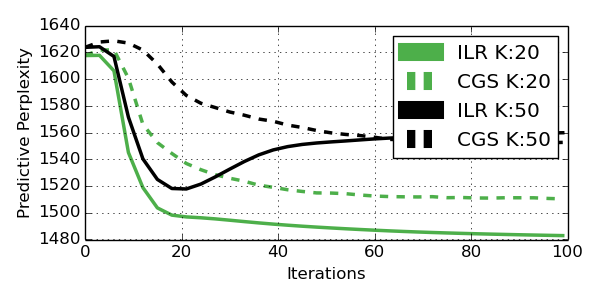}
\includegraphics[width=0.33\textwidth,height=3.7cm]{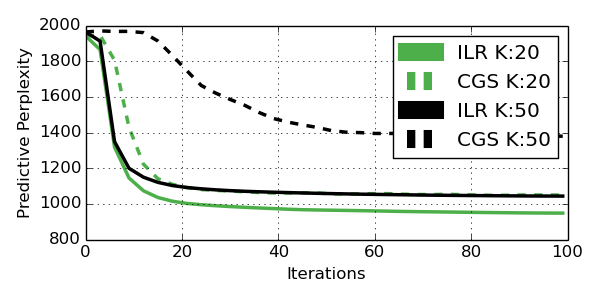}
\caption{Predictive perplexity of LDA inferred by ILR and CGS for various values of $K$ on the three datasets.}
\label{fig:lda}
\end{figure*}

\subsection{Datasets}

We used three publicly available datasets to evaluate the proposed method ILR for inference of topic models, i.e., NIPS:\footnote{\url{https://archive.ics.uci.edu/ml/datasets/NIPS+Conference+Papers+1987-2015}}\cite{NIPSDATA}, Lastfm \footnote{\url{http://www.dtic.upf.edu/~ocelma/MusicRecommendationDataset/lastfm-360K.html}}, and Movielens \footnote{\url{https://grouplens.org/datasets/movielens/}} datasets.
The NIPS dataset comes from the Neural Information Processing Systems proceedings from 1987 to 2015. 
The dataset consists of 11,040,357 records with 11,463 words, 
which are generated by 5,811 authors. 
The Lastfm dataset contains 145,534,518 records, 
which are generated based on the tuples $(user, artist, plays)$ of 360,000 users. 
Movielens is a popular benchmark dataset for movie rating prediction and recommendation. 
We used the Movielens 1M dataset that consists of 445,094 ratings generated by the most popular 1,223 users on 1,214 movies.

We conducted three types of experiments on the datasets, i.e., 
predictive perplexity,
coherence of latent topic detection,  
and sensitivity of inference to hyper-parameter setting.

\subsection{Predictive Perplexity Results}

Predictive perplexity is widely used to evaluate the generalization ability of a learning algorithm.
In this section,
we use the perplexity metric
to evaluate the proposed method ILR for inference of two well-established topic models LDA and DTM,
and compare it with the most commonly used inference algorithm collapsed Gibbs sampling (CGS).

\subsubsection{Predictive Perplexity of LDA}

                    
                     
                    


We first evaluate the performance of inference of LDA by using ILR against CGS.
In this experiment, 
the number of topics $K$ was tuned amongst $\{5,10,20,50\}$. 
The hyper-parameter $\alpha$ was set as 0.5, while $\beta$ was empirically determined to be 0.9, 0.8, and 0.5 on NIPS, Lastfm, and Movielens, respectively. 
We held out 40\% of each dataset for testing, 
and run the two inference methods for 100 iterations. 
We report the average perplexity results on five trials under the setting. 

Figure \ref{fig:lda} shows the predictive perplexity results of LDA inferred by ILR and CGS on the Lastfm, NIPS, and Movielens datasets, respectively (The lower, the better). 
Overall, the proposed method ILR (solid line) 
clearly outperforms CGS (dashed line) for inference of LDA across all the values of $K$ for each of the three datasets. 
Note that both inference methods use almost same time for running.

In addition,
there is an exception for ILR given $K=50$ on the NIPS dataset. 
As the iterations progress, 
the perplexity of LDA via ILR drops sharply, and then increases gradually.  
This is perhaps due to the choice of $K$ being too large for ILR to learn on this dataset. 
This case has been also found by Blei et al. \cite{CTM}, when they evaluated the predictive power of topic models versus the parameter $K$.
The finding may suggest that appropriate choice of hyper-parameters is advised.
In practice, we can employ cross-validation on a small development dataset to find the suitable values of the hyper-parameters.

We conclude that, given appropriate choice of hyper-parameters, 
ILR consistently outperforms CGS for inference of LDA in terms of predictive capability. 
In addition, the speed of convergence of both algorithms are comparable, however, ILR typically tends to lower bound the perplexity curves of CGS in most cases. 
We also observe that our method shows a monotonic decrease in the perplexity, and only when the curve breaks this condition, ILR will typically complete its training. This evidence may be potentially exploited as a source for hyper-parameter tuning. 
In CGS, this is not always the case, as the effect of random sampling may not satisfy the monotonic decrease of the predictive perplexity. 


\subsubsection{Predictive Perplexity of DTM}




\begin{figure*}[t]
\centering
\includegraphics[width=0.33\textwidth,height=3.7cm]{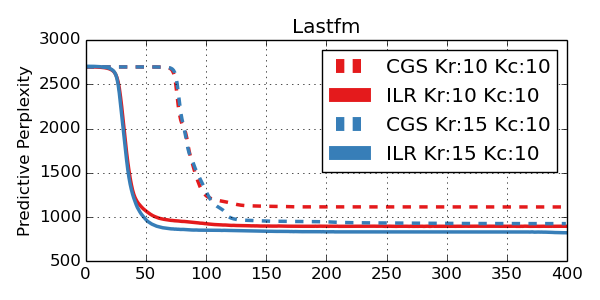}
\includegraphics[width=0.33\textwidth,height=3.7cm]{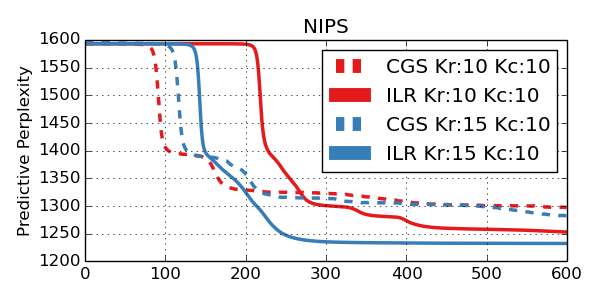}
\includegraphics[width=0.33\textwidth,height=3.7cm]{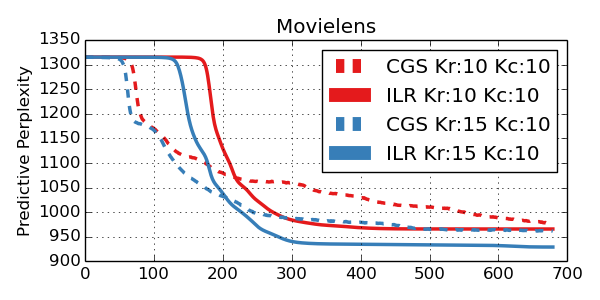}
\includegraphics[width=0.33\textwidth,height=3.7cm]{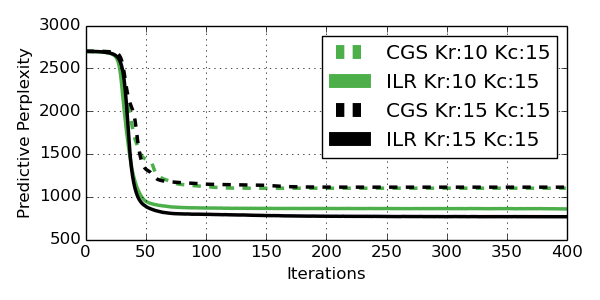}
\includegraphics[width=0.33\textwidth,height=3.7cm]{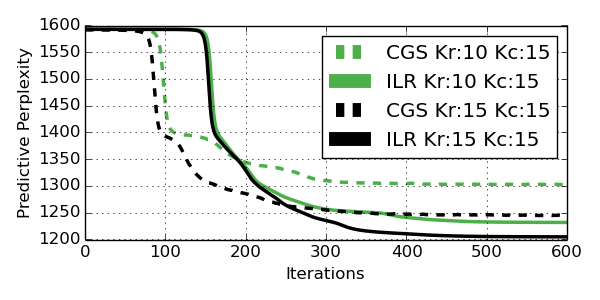}
\includegraphics[width=0.33\textwidth,height=3.7cm]{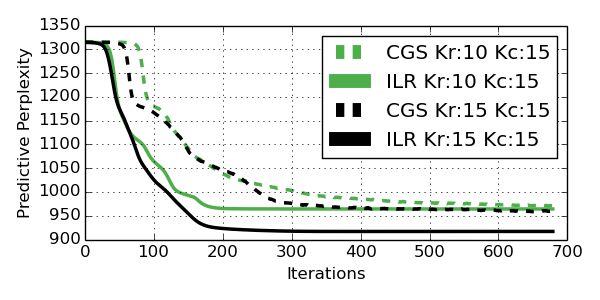}
\caption{Predictive perplexity of DTM inferred by ILR and CGS for various values of $K$ on the three datasets.}
\label{fig:dt2b}
\end{figure*}

Next, 
we further evaluate ILR against CGS for the inference of the dual topic models (DTM) using the predictive perplexity on the three benchmark datasets. 
In DTM, to estimate the probability of a document-word pair $p(w,D)$, we use the estimated parameters $\theta$, $\phi_r$, $\phi_c$ as follows:

\vspace{-0.4cm}
$$p(w,D) = \sum^{Kr}_{kr=1} \sum^{Kc}_{kc=1} p(w|\phi_{kr})p(D|\phi_{kc})p(kr,kc|\theta)$$

In this experiment, 
we run a basic grid search ranging over the $Kr$ and $Kc$ values of $10$ and $15$. 
We set the hyper-parameters $\alpha$, $\beta_r$, and $\beta_c$ as $0.5$.
Figure \ref{fig:dt2b} present the predictive perplexity of DTM as a function of the number of iterations for various values of $Kr$ and $Kc$. 
Overall, ILR again outperforms CGS for inference of the topic model DTM for various combinations of the $Kr$ and $Kc$ values on the three datasets.

It is worth noting that both inference methods ILR and CGS take a few iterations before they begin to learn. 
This is because DTM has been trying to simultaneously assign the values of two latent random variables.
We also observed that the amount of iterations needed for the methods to start learning depends on the initialization. 
A uniform initialization typically translates into more iterations taken by the methods to start learning. 
A good initialization may be to randomly recommend some parameters for $\theta$, which can be then used to generate initial topic or distribution assignments of the latent variables.

\subsection{Topic Coherence}

We use \textit{topic coherence}
to evaluate the proposed method ILR for inference of LDA in latent topic detection.
The automatic topic coherence has been well studied, and the normalized point-wise mutual information (NPMI), point-wise mutual information (PMI), and pairwise log-conditional probability (LCP) are three common metrics which have been shown to correlate positively with human judgment for topic coherence evaluation \cite{autotcoh}.
To compute the coherence of each of detected topics,
we used the co-occurrence statistics of the top \textit{N} most likely words of the topic in the corpus. 
In the experiment, 
we fixed the hyper-parameters $\alpha$ and $\beta$ to 0.5, and the number of topics $K$ to 25. 
We varied the values of $N$ ($10, 20, 50$) to study the effect of selecting different sets of the most likely words on the coherence of detected latent topics. 




\begin{table}[!t]
\caption{Automatic topic coherence results of LDA inferred by ILR and CGS, where the ILR results are highlighted in grey.}
\label{tab:topiccoherence}
  \centering
  \setlength\tabcolsep{3.0pt} 
  \bgroup
\def\arraystretch{1.2}
  \begin{tabular}{>{\scriptsize}l@>{\scriptsize}l>{\scriptsize}^l>{\footnotesize}^l>{\footnotesize}^l|>{\footnotesize}^l>{\footnotesize}^l|>{\footnotesize}^l>{\footnotesize}^l}
    \toprule
     \multirow{2}*{} &  \multirow{2}*{N} &   \multirow{2}*{} &  \multicolumn{2}{c}{\footnotesize{LCP}} &  \multicolumn{2}{c}{\footnotesize{PMI}} &  \multicolumn{2}{c}{\footnotesize{NPMI}} \\
            &  & & mean &  max , min & mean & max , min & mean & max , min \\
    \midrule
      \multirow{6}{*}{\rotatebox[origin=c]{90}{NIPS}}              
             &                \multirow{2}*{10} & CGS & -6.662 & -6.08 , -6.96 &          0.841 & 1.20 , 0.55 &          0.057 & 0.09 , 0.04 \\
             &    & \rowstyle{\cellcolor{hcolor}} ILR &  \textbf{-6.588}& -6.17 , -6.93 & \textbf{0.942} & 1.28 , 0.59 & \textbf{0.063} & 0.08 , 0.04\\
             &                \multirow{2}*{20} & CGS &  -6.682 & -6.14 , -6.92 &          0.806 & 1.03 , 0.55 &          0.054 & 0.07 , 0.04 \\
             &    & \rowstyle{\cellcolor{hcolor}} ILR &  -6.619 & -6.14 , -6.98 &          0.879 & 1.15 , 0.58 &          0.059 & 0.07 , 0.04\\
             &                \multirow{2}*{50} & CGS &  -6.687 & -6.17 , -6.86 &          0.759 & 0.91 , 0.56 &          0.051 & 0.06 , 0.04\\
             &    & \rowstyle{\cellcolor{hcolor}} ILR &  -6.641 & -6.16 , -6.93 &          0.801 & 0.98 , 0.58 &          0.054 & 0.06 , 0.04 \\
    \midrule
      \multirow{6}{*}{\rotatebox[origin=c]{90}{Lastfm}}                    
             &               \multirow{2}*{10}  & CGS & -5.031 & -4.56 , -5.79 &          1.618 & 2.54 , 0.94 &          0.128 & 0.19 , 0.08 \\
             &    & \rowstyle{\cellcolor{hcolor}} ILR &\textbf{-4.993}& -4.52 , -5.70 & \textbf{1.692} & 2.38 , 1.15 & \textbf{0.133} & 0.18 , 0.09\\
             &               \multirow{2}*{20}  & CGS & -5.205 & -4.79 , -6.02 &          1.631 & 2.49 , 1.02 &          0.125 & 0.18 , 0.08 \\
             &    &\rowstyle{\cellcolor{hcolor}}  ILR & -5.236 & -4.78 , -6.04 &          1.689 & 2.33 , 1.13 &          0.129 & 0.18 , 0.09\\
             &               \multirow{2}*{50}  & CGS & -5.572 & -5.17 , -6.32 &          1.644 & 2.36 , 1.02 &          0.119 & 0.17 , 0.07 \\
             &    &\rowstyle{\cellcolor{hcolor}}  ILR & -5.601 & -5.17 , -6.56 & \textbf{1.692} & 2.30 , 1.06 &          0.122 & 0.17 , 0.08 \\
\midrule       
     \multirow{6}{*}{\rotatebox[origin=c]{90}{Movielens}}
             &                \multirow{2}*{10} & CGS & -6.182 & -5.65 , -6.77 & \textbf{0.732} & 1.16 , 0.59 &          0.053 & 0.09 , 0.04 \\
             &    &\rowstyle{\cellcolor{hcolor}}  ILR & \textbf{-5.669} & -5.56 , -6.02 &          0.689 & 1.09 , 0.55 & \textbf{0.054} & 0.08 , 0.04 \\
             &                \multirow{2}*{20} & CGS & -6.198 & -5.74 , -6.66 &          0.724 & 1.08 , 0.63 &          0.053 & 0.08 , 0.04\\
             &    &\rowstyle{\cellcolor{hcolor}}  ILR & \textbf{-5.699} & -5.57 , -6.04 &          0.688 & 1.09 , 0.55 &          0.053 & 0.08 , 0.04 \\
             &                \multirow{2}*{50} & CGS & -6.231 & -5.78 , -6.57 &          0.723 & 1.01 , 0.63 &          0.052 & 0.07 , 0.04 \\
             &    &\rowstyle{\cellcolor{hcolor}}  ILR & -5.774 & -5.63 , -6.21 &          0.682 & 1.02 , 0.56 &          0.052 & 0.07 , 0.04\\
  \bottomrule
  \end{tabular}
  \egroup
\end{table}

Table \ref{tab:topiccoherence} shows the topic coherence results of LDA inferred by ILR and CGS in terms of LCP, PMI, and NPMI (The higher, the better).
Overall, ILR improves CGS for inferring latent topics of LDA according to the mean coherence scores. 
All the metrics show that ILR results in better performance with the exception of PMI on the Movielens dataset. 
But this is not a big issue, 
as PMI tends to assign high weights to infrequent words in a corpus, 
and is known to be not as reliable as other metrics.

The improvement of ILR over CGS remains as $N$ increases. 
The topic coherence scores of both ILR and CGS drop a little with growing the value of $N$, but this actually agrees well with expectation. 
We also observe that the mean NPMI scores are similar, for example, on the Movielens dataset, even given different values of $N$, this is perhaps due to the fact that the NPMI metric involves an additional normalization step. 




\subsection{Hyper-parameter Sensitivity}

In this section, we evaluate ILR against CGS according to the sensitivity of inference of topic models, and We compute the predictive perplexity as a function of various hyper-parameter settings. 
In this experiment, we applied CGS and ILR to the inference of LDA.
We varied the values of $\alpha$ and $\beta$ from $0.01$ to $0.5$, given fixed number of latent topics $K=25$.  
In addition, we let the methods run for 500 iterations to guarantee a fair assessment. 
Table \ref{tab:hyperparameters} presents the predictive perplexity results of LDA inferred by ILR and CGS on the Movielens dataset (The lower, the better).

\begin{table}[h]
  \caption{Sensitivity of inference of LDA to various values of hyper-parameters $\alpha$ and $\beta$ on the Movielens dataset.}
  \label{tab:hyperparameters}
  \centering
 \bgroup
 \def\arraystretch{1.2}
  \begin{tabular}{>{\footnotesize}c@>{\footnotesize}l>{\footnotesize}^l>{\footnotesize}^l>{\footnotesize}^l>{\footnotesize}^l>{\footnotesize}^l}
    \toprule
    $\alpha$ \textbackslash ~$\beta$ & Method & 0.01 &  0.05 &  0.10 &  0.25 &  0.50\\
    \midrule
	        \multirow{2}*{0.01} & CGS & 981.7 & 958.4 & 944.6 & 940.2 & 934.6\\
  & \rowstyle{\cellcolor{hcolor}} ILR & 910.7 & 906.8 & 904.7 & 901.8 & 899.7\\   
            \multirow{2}*{0.05} & CGS & 957.9 & 932.0 & 922.6 & 921.1 & 920.8\\
  & \rowstyle{\cellcolor{hcolor}} ILR & 906.8 & 904.2 & 902.6 & 900.3 & 898.3\\
            \multirow{2}*{0.10} & CGS & 945.0 & 924.8 & 920.6 & 911.2 & 910.4\\
  & \rowstyle{\cellcolor{hcolor}} ILR & 904.3 & 902.1 & 901.0 & 899.3 & 897.4\\
            \multirow{2}*{0.25} & CGS & 937.1 & 918.2 & 911.2 & 908.1 & 906.3\\
  & \rowstyle{\cellcolor{hcolor}} ILR & 900.6 & 899.0 & 898.0 & 897.3 & 895.7\\
            \multirow{2}*{0.50} & CGS & 925.1 & 914.0 & 909.3 & 904.7 & 905.2\\
  & \rowstyle{\cellcolor{hcolor}} ILR & 898.2 & 896.8 & 895.9 & 895.3 & 896.3\\
    \bottomrule
  \end{tabular}
  \egroup
\end{table}

Given the same combination of the values of $\alpha$ and $\beta$, ILR significantly improves CGS for inference of LDA. 
Surprisingly, we observe that, across all the combinations of the hyper-parameter values, the worst inference perplexity of LDA via ILR ($910.7$ when $\alpha=\beta=0.01$) is comparable with the best perplexity of LDA via CGS ($904.7$ when $\alpha=0.5, \beta=0.25$). 
Interestingly, either ILR or CGS improves the inference, as the value of the parameter $\alpha$ or $\beta$ increases.

Moreover, 
we observe that the inference of LDA via CGS is more sensitive to the choice of values of hyper-parameters.
The gap between the maximum and minimum perplexity scores is $77.0$ for CGS, while the gap is only $15.4$ for ILR, as shown in Table \ref{tab:hyperparameters}.
This agrees with expectation, as CGS is known to be sensitive to hyper-parameter settings.
In contrast, the experimental results do not show that the selection of hyper-parameters leads to the significant difference on the inference of LDA by the proposed method ILR.



\section{Conclusion}
In this paper,
we have presented an infinite latent state replication (ILR) algorithm, 
which leads to a deterministic approach to inference of topic models within Gibbs sampling framework. 
ILR benefits from the state augmentation for marginal estimation, 
and casts a given topic model to a tractable model with soft topic assignments. 
The flexibility in soft assignments results in improved generalization performance for inferring topics. 
We applied ILR to inference of two well-established topic models LDA and DTM.
Experimental results on real-world datasets validate that 
ILR outperforms CGS for the inference in terms of topic coherence and predictive perplexity, 
and the results hold despite various settings of hyper-parameters.

\vspace{-0.3cm}

\bibliographystyle{ieeetr}
\bibliography{sdm19}


\end{document}